\documentclass[fleqn,10pt]{wlscirep}
\usepackage[utf8]{inputenc}
\usepackage[T1]{fontenc}
\newcommand{\peer}{\text{peer-to-peer}}
\newcommand{\one}{\text{one-to-one}}
\newcommand{\peerData}{BeCOPE}

\usepackage{enumitem}
\usepackage{soul}
\usepackage{xcolor,colortbl}
\usepackage{multirow}
\usepackage{tcolorbox}
\usepackage{graphicx}
\usepackage{float}
\usepackage{subfig}

\usepackage{todonotes}

 \title{Critical Behavioral Traits Foster Peer Engagement in Online Mental Health Communities}

\author[1,*]{Aseem Srivastava}
\author[1]{Tanya Gupta}
\author[2,3]{Alison Cerezo}
\author[2,4]{Sarah Peregrine (Grin) Lord}
\author[1]{Md Shad Akhtar}
\author[5,6,*]{Tanmoy Chakraborty}

\affil[1]{Department of Computer Science and Engineering, Indraprastha Institute of Information Technology Delhi, New Delhi, India -- 110020}
\affil[2]{Empathy Rocks, Inc. dba mpathic.ai, Bellevue, WA 98004}
\affil[3]{Department of Counseling, Clinical and School Psychology, University of California Santa Barbara, California 93106, United States}
\affil[4]{Department of Psychiatry and Behavioral Sciences, University of Washington, 1410 NE Campus Pkwy, Seattle, WA 98195, United States}
\affil[5]{Department of Electrical Engineering, Indian Institute of Technology Delhi, New Delhi, India -- 110016}
\affil[6]{Yardi School of Artificial Intelligence, Indian Institute of Technology Delhi, New Delhi, India -- 110016}
\affil[*]{Corresponding author, Email: aseems@iiitd.ac.in, tanchak@iitd.ac.in}

\begin{abstract}
    Online Mental Health Communities (OMHCs), such as Reddit, have witnessed a surge in popularity as go-to platforms for seeking information and support in managing mental health needs. Platforms like Reddit offer immediate interactions with peers, granting users a vital space for seeking mental health assistance. However, the largely unregulated nature of these platforms introduces intricate challenges for both users and society at large. This study explores the factors that drive peer engagement within counseling threads, aiming to enhance our understanding of this critical phenomenon. We introduce BeCOPE, a novel behavior encoded Peer counseling dataset comprising over $10,118$ posts and $58,279$ comments sourced from $21$ mental health-specific subreddits. The dataset is annotated using three major fine-grained behavior labels: (a) intent, (b) criticism, and (c) readability, along with the emotion labels. Our analysis indicates the prominence of ``self-criticism'' as the most prevalent form of criticism expressed by help-seekers, accounting for a significant $43\%$ of interactions. Intriguingly, we observe that individuals who explicitly express their need for help are $18.01\%$ more likely to receive assistance compared to those who present ``surveys'' or engage in ``rants.'' Furthermore, we highlight the pivotal role of well-articulated problem descriptions, showing that superior readability effectively doubles the likelihood of receiving the sought-after support. Our study emphasizes the essential role of OMHCs in offering personalized guidance and unveils behavior-driven engagement patterns.
\end{abstract}

\begin{document}

\flushbottom
\maketitle

\thispagestyle{empty}

\noindent The prevalence of mental health distress has risen sharply in the last several years. A recent report reveals that one in six individuals suffers from mental health-related challenges\footnote{https://www.who.int/news/item/17-06-2022-who-highlights-urgent-need-to-transform-mental-health-and-mental-health-care}.
 At the same time, there is a severe shortage of mental health providers to facilitate adequate support to those in need\footnote{https://www.newamericaneconomy.org/press-release/new-study-shows-60-percent-of-u-s-counties-without-a-single-psychiatrist/} \cite{ijampaper, doi:10.1177/1178632917694350}. 
As a result of these growing challenges, we specifically examined the patterns and factors that drive individuals to engage with peer-to-peer mental health threads, focusing on the impact of behavioral, emotional, textual, and topical signals during peer-to-peer interactions. 

\begin{figure}[!t]
\includegraphics[width=\textwidth]{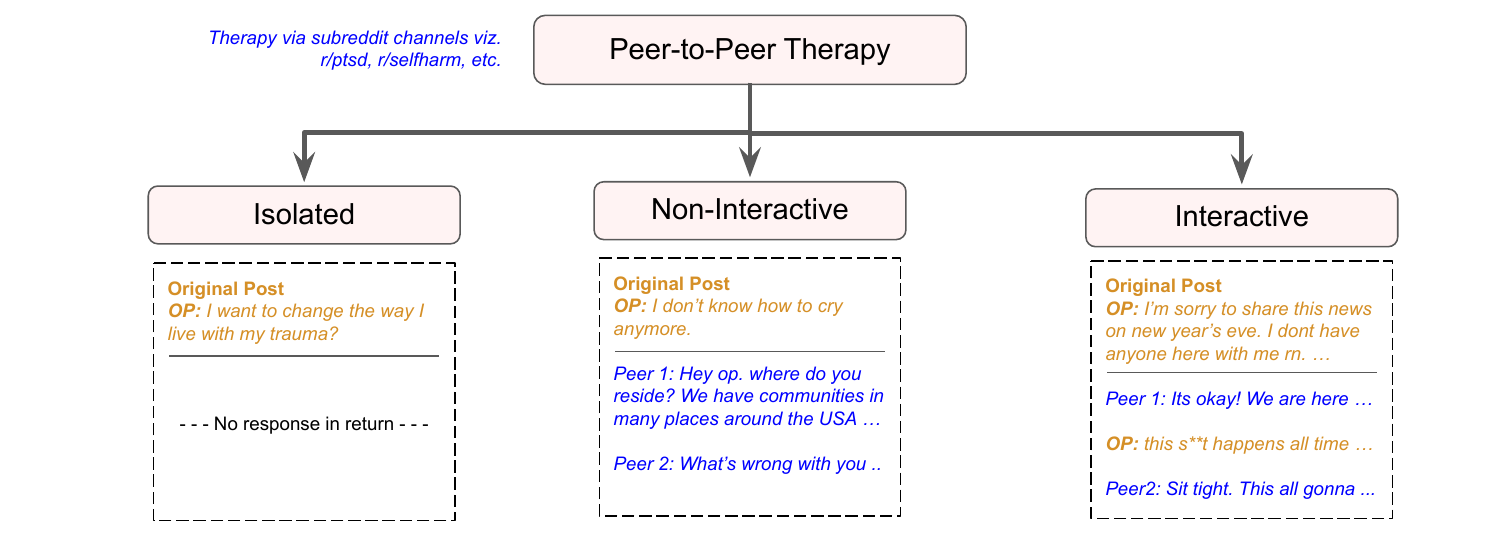}    
 \caption{{\bf Taxonomy of counseling methods along with examples. } Here, OP (original poster) is a common Internet terminology for the person who creates posts on peer-to-peer platforms. In \peer\ therapy, we inspect the level of engagement in three different categories based on the abundance of interaction with the help-seeker -- {\bf (a) interactive:} if there are back-and-forth conversations between the OP and peers, {\bf (b) non-interactive:} if the post engages peers, but the OP does not reply to peers, and {\bf (c) isolated:} if the post does not have any comment, but \one\ therapy involves the continuous exchange of dialogues between therapist and client (help-seeker).}
\label{engagement_categories}
\end{figure}

To this end, we develop the BeCOPE (BEhavior enCOded PEer Counseling) dataset, composed of peer-to-peer mental health conversational interactions across $10,118$ posts and $58,279$ comments from $21$ mental health-specific subreddits. We inspect the level of engagement on Reddit for three different OMHC categories – (a) interactive, (b) non-interactive, and (c) isolated – based on the pattern of interaction between users and the original help-seeker (see Figure \ref{engagement_categories}). Analyzing the critical factors in each engagement category, we comprehend factors and patterns that lead to constructive versus detrimental peer-to-peer mental health interactions. Understanding peer-to-peer interactions on OMHCs is key to the ethical and safe monitoring of these communities, including the moderation of safe interactions and sharing of accurate mental health information. We explore the following research questions: 
\begin{enumerate}
   \item[\bf RQ1.] When examining peer-to-peer OMHC interactions, how do intent (i.e., help-seeking), readability, and criticism impact peer willingness to engage with the original post (e.g., validation, advice-giving)?
   \item[\bf RQ2.] How does the expression of emotions in posts impact user engagement in the OMHC platforms?
\end{enumerate}

Reddit is a popular OMHC platform that has steadily emerged as a platform for seeking help concerning a spectrum of mental challenges with specific posts devoted to disorders such as depression, attention-deficit/ hyperactivity disorder (ADHD, sometimes ADD), bipolar disorder and alcohol and substance use \cite{Lokala_Srivastava_Dastidar_Chakraborty_Akhtar_Panahiazar_Sheth_2022, 10.1016/j.chb.2017.09.001, DeChoudhury2014}. Typically, users (i.e., support-seekers) create original posts to discuss their mental health issues, describing their symptoms and the contexts of their specific situations, like job loss or a recent divorce. The support-seekers, in turn, receive replies from peers (other users on the platform) with advice, recommendations for symptom management, and general support. This process allows support-seekers to share and ask for help for their mental health challenges in a cost-effective, convenient, and anonymous manner that typically results in immediate support. A recent study\cite{sharma2020engagement} analyzed patterns of posts on two popular OMHC platforms, Talklife and Reddit, by leveraging natural language processing for communication models in human-computer interaction and communication theory, operationalizing a set of four engagement indicators based on attention and interaction. The authors found that the back-and-forth peer platform communication effectively contributes to early support. A similar study\cite{10.1145/3290605.3300294} examined the change in sentiment to analyze peer-to-peer counseling settings to read whether a counseling thread or a post on the platform is correlated with a moment of cognitive change. It turned out that behavioral signals such as sentiment, affect, and topics associated with language are decisive toward effective counseling. On the same track, another study discussed the temporal engagement on social media correlating with patient disclosure\cite{Ernala}. The authors developed an autoregressive time series computational model that assesses engagement patterns and subsequently forecasts alteration in the intimacy of disclosures. They found that attributes of audience engagement, like emotional support, personal behavior, and self-disclosure, strongly predict patterns in future counseling behavior. 

Previous studies on the analysis of peer-to-peer mental health interactions identified threads that fall into affective\cite{DeChoudhury2014}, content-based \cite{Evans2012SocialSA, stephen}, and supportive \cite{10.1145/2998181.2998243} categories, thus demonstrating reliability for the functioning of peer-to-peer mental health platforms. However, little is known about how these categories of peer-to-peer mental health interactions are associated with constructive and/or detrimental outcomes. 
Understanding the characteristics of such OMHC users \cite{10.1038/s44220-023-00107-y, 10.1038/s44220-023-00085-1, 10.1038/s44220-023-00062-8, 10.1038/s44220-023-00064-6} and given the widespread use of OMHC platforms, specific patterns and factors that drive engagement in peer-to-peer mental health interaction must be identified\cite{Sharma_Choudhury_Althoff_Sharma_2020, info:doi/10.2196/mental.4418, doi:10.1176/appi.ps.201700283}. In doing so, social media platforms should be better able to monitor and intervene for the benefit of their users in distress\cite{naslund_aschbrenner_marsch_bartels_2016, https://doi.org/10.1111/jcpp.12937, https://doi.org/10.1002/eat.23148}. 

\begin{figure}[!ht]
\includegraphics[width=\textwidth]{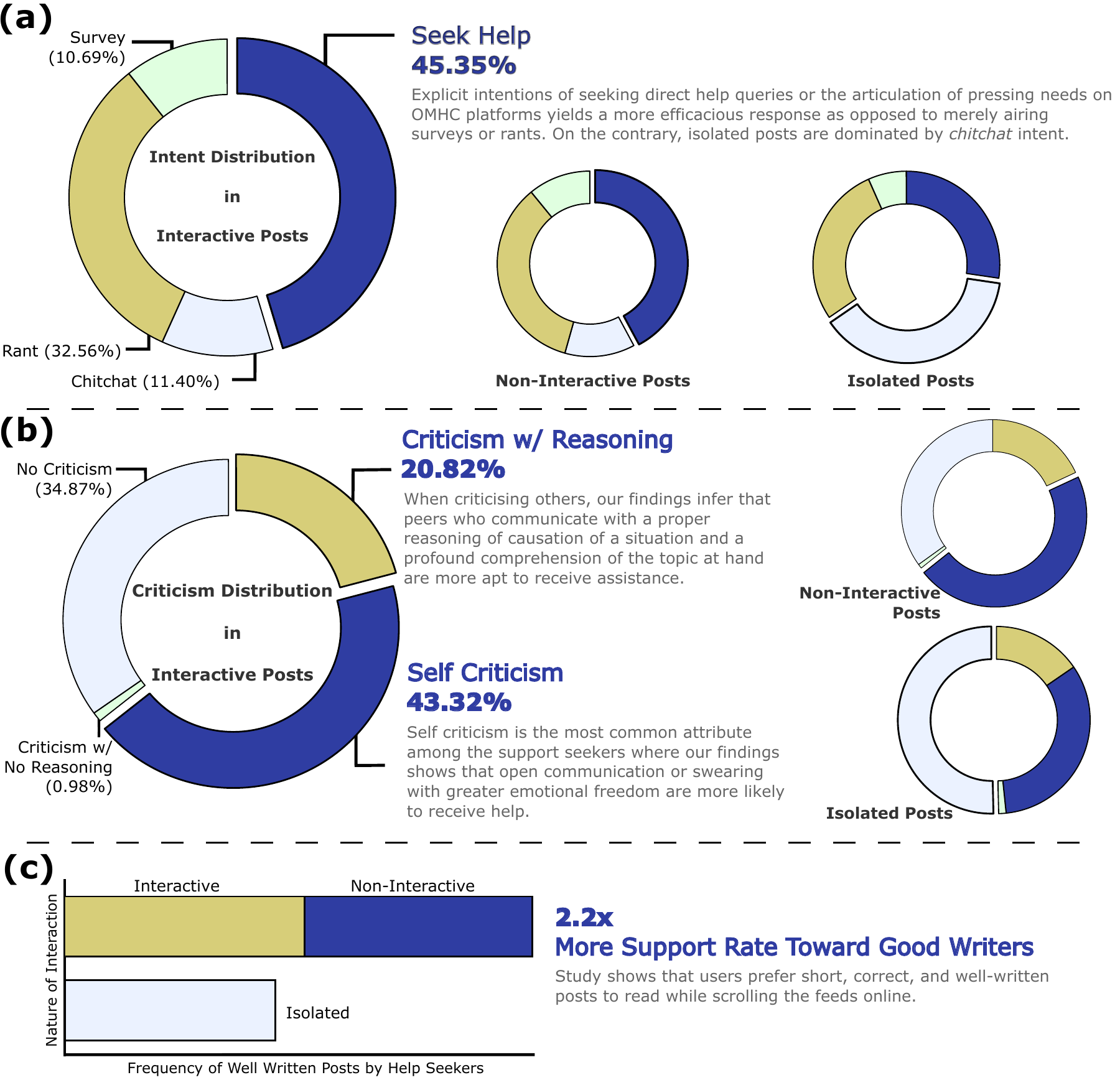}
 \caption{{\bf Distribution of behavioral signals and readability in \peerData\ across all engagement categories.} {\bf(a)} The intent distribution indicates that a majority  ($45.35\%$) of posts show explicit intentions (seek-help) through queries or the articulation of pressing needs on OMHC platforms, yielding a more productive response as opposed to merely airing surveys or rants. {\bf(b)} The criticism distribution shows that help-seekers are more likely to engage in self-criticism (43.32\%), and those who criticise openly on others with proper reasoning are more likely to receive assistance. (c) The readability statistics of posts in \peerData\ state that well-written posts receive $2.2\times$ more support (responses) as compared to poorly written posts.}
\label{fig:int_crit_read}
\end{figure}

\section*{Results}

\subsection*{RQ1: When examining peer-to-peer OMHC interactions, how do intent (i.e., help-seeking), readability, and criticism impact peer willingness to engage with the original post (e.g., validation, advice giving)?}

\paragraph{Intent.} We observe that help-seekers on OMHC platforms are $18.01\%$ more likely to receive help when they explicitly convey their pressing needs through queries, as opposed to when they make statements about their experiences. When an original post contains a help-seeking approach, it increases peer engagement. Specifically, $45.35\%$ of interactive posts, $42.16\%$ of non-interactive, and $27.34\%$ of isolated posts are help-seeking in nature, indicating that peers who explicitly ask for help for their mental issues experience greater peer engagement. We also observe that when an original post is constructed as a ``rant'' (a long statement of the problem with no explicit ask for help/advice), it receives less peer engagement. The number of isolated posts labelled with the rant intent ($38.11\%$) exceeds non-interactive posts ($34.73\%$) and interactive posts ($32.56\%$) by a margin of $3.38\%$ and $5.55\%$, respectively. Further, posts with rant intent receive the least interaction compared to other intent labels across all engagement categories, showing that the survey posts do not elicit peers’ attention toward assistance. Our analysis sheds light on RQ1 by indicating the conveyance of explicit intentions through queries or the articulation of pressing needs on the OMHC platforms yields a more efficacious response. We present the distribution of intents across three engagement categories in Figure \ref{fig:int_crit_read}{\color{blue}{(a)}}.  The four annotated intent labels receive a significant agreement score with a confidence $\geq 95\%$ on the $p$-values of help-seeking ($0.022$), rant ($0.046$), chitchat ($0.016$), and survey ($0.028$). Furthermore, {\color{blue}{\em SI Appendix} (Section 1)} presents fine-grained details of intent labels and their annotation.

\paragraph{Criticism.} We observe that isolated posts have maximum \textit{no-criticism} (NC) labels (50.34\%) as compared to non-interactive (34.92\%) and interactive (34.87\%) posts. Figure \ref{fig:int_crit_read}{\color{blue}{(b)}} shows the distribution of {\em criticism} labels across all engagement categories. Conversely, individuals who can obtain support from their peers on OMHCs are frequently found to engage in criticising themselves and others. We bifurcate the criticism of others into two indicative categories -- \textit{criticism with reasoning} (CR) (i.e., a logical presentation of one's experience), and \textit{criticism with no-reasoning} (CNR). Out of all three engagement categories, interactive engagement carries the maximum CR label, $2.75\%$ and $5.39\%$ more than non-interactive and isolated engagement categories, respectively. This trend directly draws attention to the fact that proper reasoning in criticism is vital for receiving help. In contrast, CNR is most prevalent in the isolated engagement, highlighting that criticism without proper reasoning only adds noisy understanding to the reader's mind. Similarly, {\em self-criticism} is considered the most prevalent type of criticism  among those who receive help. This implies that people seeking support are more likely to engage in self-criticism, and those who express their emotions more openly are more likely to receive assistance. As a result, we infer that peers who criticise and have a profound comprehension of the topic at hand are more apt to receive assistance.  {The four annotated criticism labels receive adequate agreement score with confidence $\geq 95\%$ on the $p$-values of {\em criticism w/ reasoning} ($0.043$), {\em criticism w/ no reasoning} ($0.010$), {\em no criticism} ($0.009$), and {\em self-criticism} ($0.035$). Additional details related to the selection and annotation of the criticism labels are presented in {\color{blue}{\em SI Appendix} (Section 1)}.

\paragraph{Readability.} We hypothesize that well-written posts (i.e., easier to read) foster better understanding and subsequently attracted more peers to engage. Our initial observation supports the hypothesis; most of the posts in the BeCOPE dataset are hard to read, i.e., rated $\leq 2$ on a scale of $1$ to $5$, with $1$ being the least comprehensible. Our analyses reveal that posts scoring higher in readability result in $2.2\times$ greater support ratings from peers, as shown in Figure \ref{fig:int_crit_read}{\color{blue}{(c)}}. We further employ experts in linguistics to understand what contributes more toward understanding posts. We observe that factors like the length of the post, the division into paragraphs and listicles, grammar, spelling, clarity of the issue, and usage of short forms (SMS language) are critical that peers take into consideration when reading and deciding to engage with a post. {The readability score receives significant confidence of $\geq 95\%$ with average $p$-values across all five labels to be $0.040$. More details related to the annotation of readability score are presented in {\color{blue}{\em SI Appendix} (Section 1)}.

\subsection*{RQ2: How does the expression of emotions in posts impact user engagement in the OMHC platforms?}

\paragraph{Emotion labels.} Emotions play a vital role in mental health support seeking. Empathetic understanding is an attempt by the observers/experts to regulate emotions that help-seekers express\cite{kim2015understanding}. Figure \ref{fig:emotion_topic} shows a frequency-based radial distribution of the most frequent emotion labels in \peerData. Our analysis of emotion labels shows that $10\%$ of the isolated posts carry {\em neutral} emotion labels. In contrast, only $3\%$ posts carry {\em neutral} emotions for both interactive and non-interactive posts combined. Furthermore, $12.3\%$ of the non-isolated posts exhibit {\em curiosity} as the secondary emotion compared to $7\%$ isolated posts. Evidently, labels such as {\em sadness, curiosity, fear,} and {\em realization} are more prevalent in non-isolated posts. On the other hand, emotion labels such as {\em caring, confusion, approval, joy,} and {\em neutral} are more prevalent in isolated posts. Consequently, peers exhibiting explicit emotional expression in posts, such as curiosity, fear, and sadness, receive more significant support in $86\%$ of the cases. For the remaining $14\%$ of the posts,  emotions are observed to be with tepid emotional labels, such as caring, confusion, or neutral, to which peers often ignored responding, leading to no interaction. 

On analyzing a sample of 100 posts, we subjectively categorize extreme emotions expressed into various types, including fear, excitement, sadness, etc. In the category-wise emotion distribution (Figure \ref{fig:emotion_topic}), we observe that posts expressing such explicit extreme emotions have a higher chance of receiving a response, whereas posts with tepid emotional labels, such as caring, confusion, and neutral tend to be ignored.

\begin{figure}[!t]
\includegraphics[width=\textwidth]{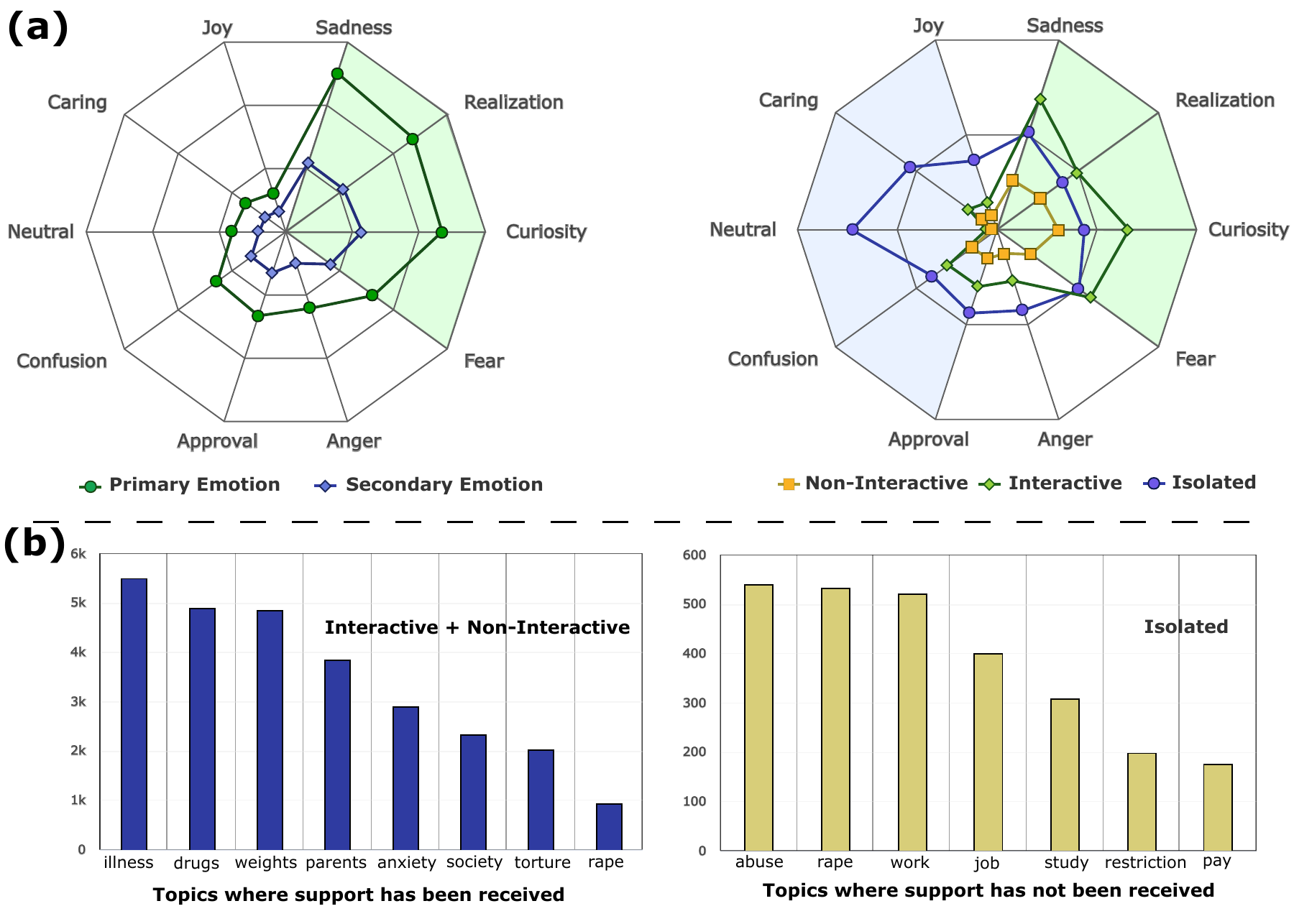}
 \caption{{\bf (a)} {\bf Distribution of emotion labels in the \peerData\ dataset. }For brevity, we show plots for the top 10 emotion labels only. Each post is tagged with primary and secondary emotion labels. We further analyze the emotion label distribution across three engagement categories. {\bf (b) Topical analysis on the \peerData\ dataset.} We perform Latent Dirichlet Allocation (LDA) \cite{blei2003latent} to form $8$ clusters of topics. To analyze the topics on which peers respond, we club interactive and non-interactive posts, where peers respond and compare them with topics from isolated posts.
 }
\label{fig:emotion_topic}
\end{figure}
 
\subsection*{Metadata and Content Analysis}
We conduct an auxiliary analysis of the BeCOPE dataset with a prime focus on metadata and textual properties. These experiments aim to assess the impact of minor actions, such as subjectivity, interaction count, time of posting, anonymity, etc., on help-seeking. We conclude that specific minor actions taken by help-seekers on OMHC platforms can increase the probability of receiving assistance. Our initial findings suggest that descriptive titles and body content attract more help than compact usage of words. Likewise, the active participation of the help-seeker in the conversation (through comments) increases the chances of receiving help two-fold. Such approaches might assist help-seekers in gaining early access to assistance. Similarly, the time of seeking help also plays a vital role in peer assistance to the help-seeker. To this end, we further extend our analyses to understand the impact of time of seeking help on OMHC platforms. The results show that -- first, seeking help during the night hours is the most common time to post; second, the availability of helpers is also at a maximum during night hours. These findings are essential because they reveal that many users seek help on OMHCs when health providers are typically unavailable, particularly during business hours. It is, therefore, critical that help-seekers post to the most suitable mental health subreddit in their time of need. We observe that a few mental health subreddit channels like r/OpiatesRecovery are entirely dedicated to providing frequent assistance to help-seekers, including during late hours. A detailed analysis with additional experiments is presented in { \color{blue}{\em SI Appendix} (Section 2)}.

\subsection*{Topical Analysis}
We also perform a topical analysis of peer-to-peer interactions, aiming to understand what specific topics and keywords drive the conversation in three engagement categories (viz. interactive, non-interactive, and isolated). To this end, we apply Latent Dirichlet Allocation (LDA)\cite{blei2003latent} on the posts in each engagement category. The idea is to understand the topics on which peers respond and don't respond. Therefore, we segregate isolated and non-isolated posts to study the topics on which the support is received and not received, respectively. We observe that the most ordinary topics are in isolated posts, which include discussions about school-related issues, abuse, rape, pressure to meet society’s standards, salary, and freedom to express opinions and feelings. On the other hand, we observe that the frequently discussed topics from the non-isolated category are anxiety, drugs, common symptoms/illness and diagnosis, parenting behaviors, body image issues, food and weight, anxiety, and relapsing on drugs. 
Figure \ref{fig:emotion_topic} shows a cluster of topics for posts from each category to obtain the most common topics in conversations. Evidently, the common topics of discussion in isolated posts elucidate that people shared experiences about many sensitive and stigmatized issues; subsequently, they remain unexplored, as indicated by the number of isolated posts. As a result, ordinary topics that resonate with peers and enjoy widespread prevalence tend to attract more interactions and are more likely to receive active engagement from peers on OMHCs.

\section*{Discussion}
Understanding user behavior and online engagement is consistently challenging, particularly in comprehending the complexities of individuals in distress. OMHC platforms have emerged as crucial spaces for peer-based mental health discussions, enabling individuals to discuss their intrinsic thoughts and mental health issues openly. Beyond the OMHC's function, only a handful of these users interact, with even fewer users receiving the anticipated assistance. 
The most effective way of assessing peer engagement is to understand the factors on which peer interaction depends. 
Platforms like Reddit, containing dedicated mental health subreddits, offer rich repositories of discussions on relevant topics. Our formulated hypothesis posits that the comprehension of peer behavioral attributes such as intent, criticism, and readability significantly contributes to a holistic understanding. In addition, the expressivity of emotions on OMHCs can further concentrate on the causal underpinnings of these behavioral dynamics. However, this research area has remained under-resourced and insufficiently explored. Our newly introduced \peerData\ dataset holds significant implications beyond the insights drawn in this study. It can serve as a valuable resource across various research domains with dimensions ranging from empathetic to behavioral conduct of peers on OMHCs and further epitomizing explanations and casualty of such implicit underlying causes.

Our research examines the behavioral, emotional, and topical dynamics associated with varying levels of engagement among peers within OMHCs. We perceive engagement as an indication of a peer's preparedness to provide support. Our findings underscore that simple behavioral characteristics such as explicitly seeking help and refraining from criticizing others can increase peer engagement, as observed in $\sim$$50\%$ of the cases. This observation emphasizes that behaviors like ranting, criticising others, and generic chit-chatting do not elicit productive peer attention.
At the same time, users express themselves in different styles, and the underlying concept of peers being able to understand others hinges on the clarity of the posts' readability. Earlier research shows that using short sentences is more engaging \cite{10.1162}. In contrast, we show that peers with intricate thoughts aren't constrained to concise posts; instead, they often require more extensive elaboration \cite{doi:10.1177/0044118X221129642}. Our research demonstrates a twofold increase in support for individuals openly expressing their concerns on the OMHC platforms.
Conversely, the illustration of emotion dynamics is an additional gauge to evaluate the user's context. In alignment with our formulated hypothesis, the intricate interplay of emotions articulated within OMHCs demonstrated a direct correlation with the level of peer interaction. Analogous to socio-cultural implications, instances where individuals convey heightened emotional intensity consistently involve more engagement, while expressions characterized by emotional neutrality tend to diminish in terms of peer involvement. This phenomenon potentially stems from underlying factors such as relatability, the emergence of a palpable sense of urgency, and a compelling inclination to provide empathetic validation and support. These emotionally charged interactions establish a conspicuously relatable presence, effectively motivating peers to participate in discussions and disseminate adaptive coping techniques actively.
Consequently, the assessment of peer engagement within OMHCs stands as pertinent societal research that aims to assess the intricate dynamics underpinning an effective peer support framework. Such OMHCs serve as forums where peers engage in a wide spectrum of discussions, yet only a few receive the required assistance. We are convinced that a crucial void in this landscape lies in fostering societal awareness regarding the nature of these challenges and their appropriate navigation. For instance, individuals often discuss sensitive and stigmatized matters, which, although prevalent in volume, remain relatively unexplored, as substantiated by the prevalence of isolated posts. As a result, topics of a more general nature are observed to attract increased interaction. Furthermore, there exist a few impactful takeaways from our auxiliary content (metadata) analysis. We present a detailed discussion in {\color{blue}{\em SI Appendix} (Section 3)}. These perceptive insights inherently underscore the significance of understanding the factors of the support ecosystem before its effective utilization for constructive engagement. 

\section*{Conclusion}
OMHC platforms have become a popular way to seek help for people struggling with mental health issues \cite{10.1371/journal.pone.0053244, 10.1371/journal.pone.0110171, https://doi.org/10.1111/jcpp.13190, SocialMediaAdolescents}. Our work analyzed the granular user posting behaviors that foster peer engagement with the mental health content on OMHC platforms, specifically subreddits. The primary aim of this work was to better understand the behaviors of support seekers and the factors that drive peer engagement with the original post. We found that the intent of a post (seeking support versus ranting about one's experience), the readability, and the criticism elements of a post were associated with peer engagement. Further, we also found that emotional expression, the original post's content, and contextual details like the time that a post was made impacted peer engagement. Our proposed dataset and empirical study call for more research to understand peer engagement on mental health platforms, including elements that lead to constructive versus detrimental engagement\cite{https://doi.org/10.1111/jcpp.13190, https://doi.org/10.1002/jclp.23486, Torous116}. These data are critical in understanding how OMHC can best support users experiencing distress in addition to preventing the proliferation of harmful and inaccurate mental health advice and information\cite{palmeretal, 9018280, 9115602}. 

Understanding user behavior and online activity is challenging, and even harder to understand individuals in distress. The current study primarily focused on \peer\ engagement concerning mental health content. We understand that the findings can vary across other platforms like Twitter, Talklife, 7Cups, Facebook, Instagram,  and even other subreddit channels. 
The future direction of this work will be to better understand user behavior on OMHCs, including how to monitor and moderate peer engagement so that it is not harmful to individuals in distress. Although our findings shed light on the connecting patterns of \peer\ online engagement, more research is needed to develop computational methods to gauge user satisfaction and behavior by exploiting the annotations we have done in \peerData.

\section*{Methods}
\subsection*{Data Collection}
To study latent signals in peer-to-peer mental health interactions, we develop BeCOPE by curating posts from $21$ subreddits. Reddit is organized into spaces called subreddits, where each subreddit is specific to a certain discussion topic. To analyze behaviors on peer-to-peer mental health platforms, we scraped, processed, and annotated subreddit data to develop the dataset. We explored numerous subreddits and handpicked  $21$ most active mental health-related subreddits, as shown in Table \ref{tab:peerDataFullStats}. For each shown subreddit, we curated $500$ posts and their comments from January 2020 to December 2020. Further, we performed a sanity check to ensure that conversations were acceptable (e.g., noise-free, written in English). We collected $10,118$ posts and $58,279$ comments along with their metadata, such as author information, score (upvotes), time of creation, and the number of comments.

\begin{table}[!t]\centering
\footnotesize
\resizebox{\textwidth}{!}{%
\begin{tabular}{l|cc|cccc|cccc|cc}
\toprule
\multirow{3}{*}{\bf Subreddits} & \multirow{3}{*}{\bf Posts} & \multirow{3}{*}{\bf Comments} & \multicolumn{4}{c|}{\bf Intent} & \multicolumn{4}{c|}{\bf Criticism} & \multicolumn{2}{c}{\bf Readability}\\
&&& \multirow{2}{4em}{\centering \textbf{Help seeking}} & \multirow{2}{*}{\textbf{Rant}} & \multirow{2}{*}{\textbf{Survey}} & \multirow{2}{*}{\textbf{Chitchat}} & 	\multirow{2}{4em}{\centering \textbf{Self criticism}} & 	\multirow{2}{4em}{\centering \textbf{Other w/ Reason}}  & 	\multirow{2}{5em}{\centering \textbf{Other w/o Reason}}& 	\multirow{2}{4em}{\centering \textbf{No criticism}} & \multirow{2}{*}{\textbf{Clear}}  & \multirow{2}{*}{\textbf{Non-clear}} \\
& & & & & & & & & & & & \\
\cmidrule{1-13}

r/Anxiety &	469 & 1773 & 252 & 129 & 62 & 26 & 278 & 48 & 7 & 136 & 467 & 2\\
r/ptsd	&	494	&	1567	&	221	&	144	&	64	&	65	&	180	&	135	&	1	&	178	&	494	&	0\\
r/suicideWatch	&	403	&	2545	&	90	&	246	&	17	&	50	&	231	&	34	&	10	&	128	&	378	&	25\\
r/addiction	&	487	&	3581	&	217	&	148	&	43	&	79	&	246	&	67	&	6	&	168	&	466	&	21\\
r/ADHD	&	423	&	3856	&	169	&	104	&	78	&	72	&	139	&	31	&	9	&	247	&	418	&	5\\
r/alcoholicsanonymous	&	498	&	6021	&	181	&	107	&	47	&	163	&	155	&	58	&	5	&	280	&	490	&	8\\
r/Anger	&	464	&	2620	&	233	&	184	&	31	&	16	&	245	&	140	&	16	&	63	&	462	&	2\\
r/BPD	&	519	&	2744	&	180	&	185	&	113	&	41	&	234	&	99	&	4	&	182	&	518	&	1\\
r/depression	&	547	&	1951	&	83	&	363	&	26	&	75	&	243	&	91	&	18	&	195	&	546	&	1\\
r/domesticviolence	&	425	&	2847	&	254	&	94	&	25	&	52	&	34	&	277	&	1	&	113	&	421	&	4\\
r/eating\_disorders	&	568	&	2021	&	256	&	209	&	51	&	52	&	346	&	43	&	1	&	178	&	567	&	1\\
r/getting\_over\_it	&	476	&	2551	&	230	&	163	&	35	&	48	&	258	&	72	&	2	&	144	&	473	&	3\\
r/mentalillness	&	484	&	1895	&	208	&	155	&	52	&	69	&	209	&	99	&	2	&	174	&	480	&	4\\
r/OpiatesRecovery	&	493	&	6112	&	215	&	116	&	62	&	100	&	185	&	28	&	3	&	277	&	493	&	0\\
r/rapecounseling	&	481	&	2390	&	288	&	142	&	26	&	25	&	125	&	269	&	1	&	86	&	481	&	0\\
r/sad	&	486	&	2258	&	44	&	287	&	27	&	128	&	115	&	71	&	8	&	292	&	485	&	1\\
r/selfharm	&	467	&	1928	&	136	&	232	&	52	&	47	&	243	&	39	&	0	&	185	&	465	&	2\\
r/selfhelp	&	419	&	2001	&	177	&	60	&	28	&	154	&	163	&	37	&	0	&	219	&	390	&	29\\
r/socialanxiety	&	461	&	2798	&	167	&	128	&	64	&	102	&	201	&	58	&	0	&	202	&	428	&	33\\

r/OCD	&	424	&	2528	&	159	&	117	&	63	&	85	&	209	&	29	&	3	&	183	&	424	&	0\\
r/helpmecope	&	473	&	2121	&	277	&	127	&	17	&	52	&	170	&	160	&	2	&	141	&	471	&	2\\

\toprule

\bf Total	&	\bf9961	&	\bf58108	&	\bf4037	&	\bf3440	&	\bf983	&	\bf1501	&	\bf4209	&	\bf1885	&	\bf99	&	\bf3771	&	\bf9817	&	\bf144\\

\cmidrule{1-13}
\rowcolor{yellow!20}
\bf IAA	($\kappa$)&-&-&	\multicolumn{4}{c|}{\bf 0.963} & \multicolumn{4}{c|}{\bf 0.888}& \multicolumn{2}{c}{\bf 0.861}\\
\bottomrule

\end{tabular}}
\caption{{\bf Statistics of the \peerData\ dataset}. We collected a total of $\sim10K$ posts and $\sim50K$ comments. We annotated all the posts using three core labels -- (i) intent, (ii) criticism, and (iii) readability (Clear: Excellent, Good and Average; Non-clear: Mediocre and Poor). IAA ($\kappa$) represents the inter-Annotator agreement using Cohen's kappa score.}
\label{tab:peerDataFullStats}
\end{table}

\paragraph{\bf Step 1: Categorization of interactions by the level of peer engagement.}
Depending on the comments on a post, we classified the collected conversations into one of the three engagement categories: (i) interactive, (ii) non-interactive, or (iii) isolated. If an original post involved back-and-forth comments from the original user and peers, the conversation was deemed “interactive” (see {\color{blue}{\em SI Appendix}, Table S1} for an example). If an original post had zero comments, the conversation was deemed “isolated.” Finally, if an original post received more than one comment from peers, but the original user did not acknowledge or reply to peers’ comments, the conversation was deemed “non-interactive”. 

\paragraph{\bf Step 2: Annotation of posts by behavioral and emotional labels.} The first step in the annotation process was the curation of Reddit posts on mental health topics by categorizing them based on (i) intent, (ii) criticism, (iii) readability, and (iv) emotion labels. We manually annotated $\sim$5K posts and subsequently learned respective classifiers to obtain pseudo-labels for another $\sim$5K posts. Next, a sanity check of the annotated dataset was performed to ensure the reliability of the annotations. Finally, we used the resultant dataset of $\sim$10 posts for our analyses. Detailed statistics of the annotated BeCOPE dataset (including pseudo labels, discussed later) are presented in Table \ref{tab:peerDataFullStats}. We discuss the pseudo-modeling and annotation details in {\color{blue}{\em SI Appendix} (Section 1: Emotion)}.

\begin{figure}[!t]
\includegraphics[width=\textwidth]{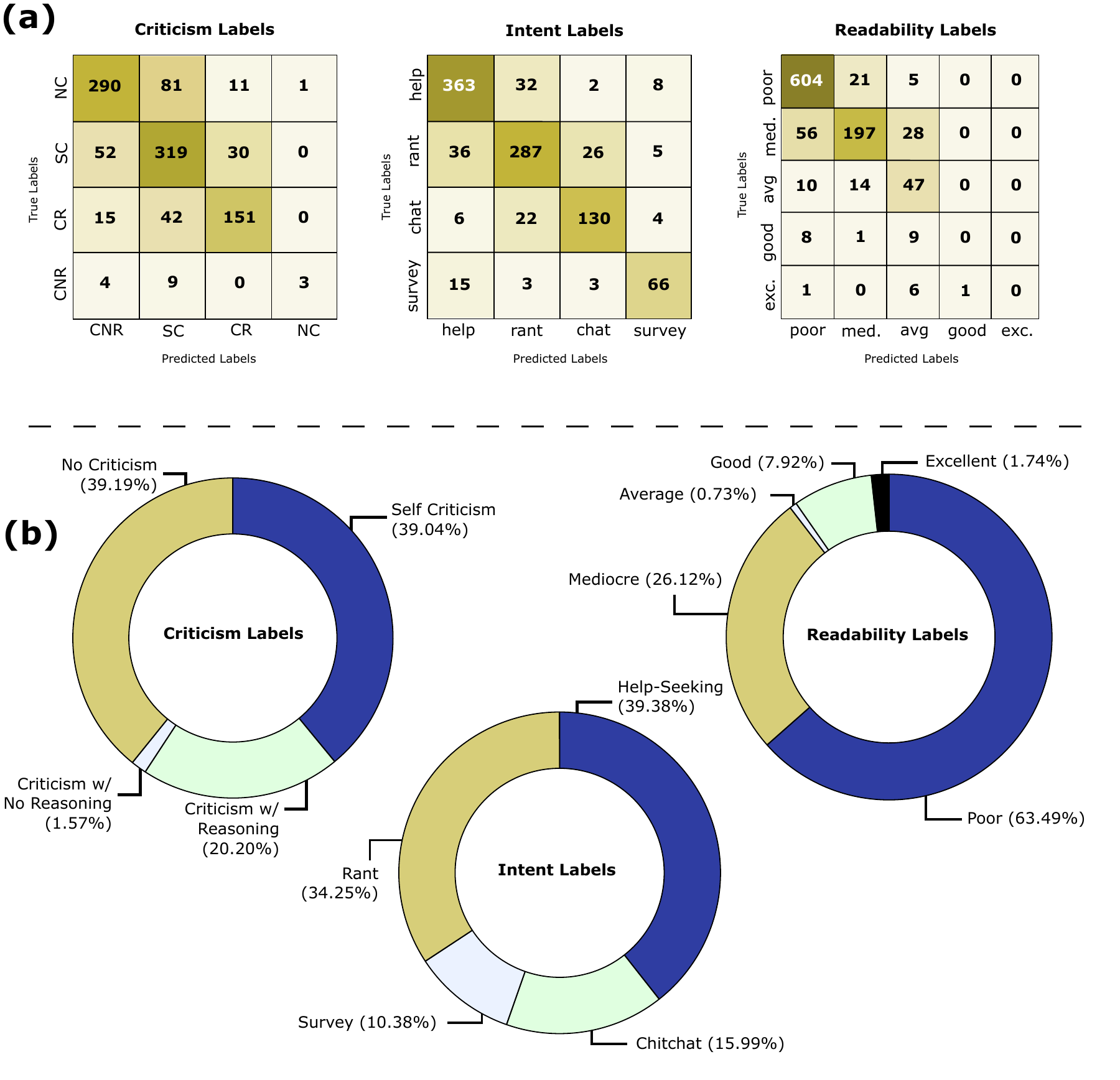}    
 \caption{{\bf (a)} Confusion matrix to represent the performance of pseudo labeling of criticism, intent, and readability labels. We exploit BERT to fine-tune on \textasciitilde5K manually annotated posts to predict criticism, intent, and readability on the remaining posts. {\bf (b)} Distribution of behavioral signals (criticism and intent) along with readability in the complete \peerData\ dataset.}
\label{data_annotation_label}
\end{figure}

\subsection*{Ethical Consideration} 
Considering the sensitivity of research in mental health, this paper does not include any personal, identifiable information of any OMHC user. Further, our models involve sophisticated deep-learning models, which are careful not to take any bias toward any gender, caste, race, diagnosis, or peers with specific symptoms. We collected data solely based on the most relevant mental health subreddits and did not include any bias in the choice of particular subreddit channels. Finally, we conducted all experiments without compromising the anonymity of online users in \peerData.

\section*{Data Availability}
The  \peerData\ data used for this study's analysis has undergone a rigorous pipeline and subjected to expert validation to ensure its quality and relevance. Researchers interested in utilizing this dataset for their research purposes can request access by contacting the corresponding authors. A sample of the \peerData\ dataset is available at \url{https://github.com/LCS2-IIITD/peer_study_omhc}. This sample dataset provides an overview of the type and structure of the data and can aid reviewers in understanding the scope and nature of the dataset used in this study.

\section*{Author Contributions}
A.S., M.S.A., and T.C. conceived and designed the study. A.S. and T.G. performed the experiments. A.S., T.G., M.S.A., and T.C. acquired, analyzed and interpreted the results. All the authors drafted the paper. A.S., M.S.A., A.C., S.P.L., and T.C. critically revised the paper. M.S.A. and T.C. supervised the work. T.C. arranged the funding.

\section*{Funding Information}
The work is financially supported by ihub-Anubhuti-iiitd Foundation, set up under the NM-ICPS scheme of the DST.

\section*{Competing Interests}
The authors declare no competing interests.

\section*{Additional Information}
{\bf Supplementary information.} The online version contains supplementary material.\\

\noindent{\bf Correspondence and requests for materials} should be emailed to Aseem Srivastata (aseems@iiitd.ac.in) and Tanmoy Chakraborty (tanchak@iitd.ac.in).

\bibliography{main}

\clearpage
\begin{center}
{\huge Supplementary Information}    
\end{center}

\section{Data Annotation} 
Peer-to-peer counseling conversations are open-ended, where users express their diverse and different perspectives. We observe that users exhibit a variety of intents while discussing mental health issues \textit{viz.} rating, seeking help survey, or doing general chit-chat. On the other hand, some alleged criticism for their issues. 
Moreover, another aspect of  \peer\ counseling is to understand whether users convey their needs clearly and crisply. A well-written post may have attracted more and perhaps better responses than a poorly-written post. Therefore, the readability of the post is another factor that directly affects the chances of receiving help on a mental health post from peers. 
At the same time, studying the expressed emotions is highly imperative. We hypothesize that all the above-discussed factors -- the knowledge of intent, the presence of criticism, the study of emotions, and the readability of the posts, are crucial in understanding the need of the help-seeker and accordingly providing appropriate assistance. 
Considering the literature and observations, we designed a set of guidelines to annotate the curated Reddit posts. A detailed discussion of the guidelines considering the four factors is presented below. 

\subsubsection*{Intent} Intent defines the purpose of the original poster (OP) in the post. We divide the posts into four categories based on the user's needs: help-seeking, rant, survey, and chitchat.
    \begin{itemize}[leftmargin=*]
        \item \underline{Help-Seeking:} Original posters explain mental health issues and expect peers to provide helpful suggestions to improve their condition.
        \item \underline{Rant:} Original posters share their (strong) views on mental health issues without expecting help from peers.
        \item \underline{Survey:} Original posters share mental health issues and ask peers to share their experiences. \textit{Survey} differs from \textit{help-seeking} as survey-labeled posts ask for a generic point of view on related mental health issues rather than individual-centric assistance. 
        \item \underline{Chitchat:} The Chitchat label is used for filler posts that are not directly related to mental health issues. Such posts include well wishes, general guidelines, occasional greetings, etc. 
    \end{itemize}
    \label{criticism_data}
    
\subsubsection*{Criticism} Original posters often criticize the situation caused due to their or others' mental health issues. Sometimes,  criticism is on their own; other times, it is on others. Hence, it is important to study if showing criticism could be a cause to receive better help. In other words, do peers prefer helping others who use criticizing language in posting their mental state? To understand this, we define four criticism labels: \textit{no-criticism, self-criticism, others' criticism with reason} and \textit{without reason}. 
    
    \begin{itemize}[leftmargin=*]
        \item \underline{Self-Criticism (SC).} We use this label for posts where original posters criticize themselves for their mental health issues.
        \item \underline{Criticism on Others with Reason (CR).} We use this label for posts where original posters criticize others for their mental health issues. Also, they provide some reasons (justification) to support their criticism. 
        \item \underline{Criticism on Others with No Reason (CNR).} This label differs from CR as the criticism is not backed by reasoning.   
        \item \underline{No-Criticism (NC).} We use this label for posts where there is no criticism. 
    \end{itemize}

\subsubsection*{Readability}  
Readability is essential in imparting most of the information via textual communication in all professional domains\cite{BONSALL2017329}. Earlier works utilized the readability criteria to decide the impact of a mental health post using statistical properties of posts such as length of the post \cite{sharma2020engagement}. However, we argue that a shorter post could also be interpreted as poorly readable. Therefore, in this work, we define the readability score based on the clarity of the text and the amount of effort one needs to put into comprehending the post. We observe that lengthier sentences pose a degree of uneasiness in readers besides the use of SMS slang and abbreviations. Based on our observations, we define five readability levels for a post -- \textit{excellent}, \textit{good}, \textit{average}, \textit{mediocre}, and \textit{poor}. 
    
\subsubsection*{Emotion}  Emotion labeling is the practice of cultivating empathetic knowledge in conversations. We employ a set of 28 emotion classes --\textit{admiration, amusement, anger, annoyance, approval, caring, confusion, curiosity, desire, disappointment, disapproval, disgust, embarrassment, excitement, fear, gratitude, grief, joy, love, nervousness, optimism, pride, realization, relief, remorse, sadness, surprise}, and \textit{neutral}-- for our Reddit posts. Moreover, we observed that many posts conveyed multiple emotions in a single post; hence, we assigned two emotions for each post, i.e., the \textit{primary emotion} and the \textit{secondary emotion}.

\begin{table}[ht]\centering
\footnotesize
\resizebox{\textwidth}{!}{%
\begin{tabular}{p{40em}|c|c|c|c|c}\toprule

\rowcolor{cyan!20} & & & \multicolumn{2}{c|}{\bf Emotion} &  \\
\rowcolor{cyan!20} \multirow{-2}{*}{\bf Post} & \multirow{-2}{*}{\bf Intent} & \multirow{-2}{*}{\bf Criticism} & \bf Primary & \bf Secondary & \multirow{-2}{*}{\bf Read}\\
\cmidrule{1-6}
Hello all. I unfortunately used again today, despite going to my first NA meeting last night. My loneliness, is a trigger. I spent the whole day by myself, and decided it wouldn't hurt to light up one more time. I have a job interview, next Monday. I don't want to lose my life. I am usually a proud person, and solve my problems on my own. But, I know I need to reach out to someone before I spiral out of control. I honestly see myself giving up everything for it. Please, any advice and tips would help me in this low moment. I have no one in my life currently. Thank you. [\colorbox{green}{\textbf{Interactive}}] &   \textit{Help-seeking}    &   SC  &   \textit{Sadness} &   \textit{Gratitude}   &  \textit{Excellent} \\
\cmidrule{1-6}
I recently found out my boyfriend has been crushing or opening then snorting most of his prescription medications. We had been arguing non-stop over his marijuana addiction, but this brings it to the next level.Not just his ADHD meds, but his depression and anxiety meds too. He claims it makes them more effective. I tried negotiating with him over letting me dispense his ADHD meds to him one week at a time, and he goes on tirades about loss of control/lack of trust/I’m not his mother/the government and his doctors can’t control him etc. I already am exhausted from our constant fighting over marijuana abuse.He says it’s not my problem because it’s his body and his decision. He has ADHD, depression, and anxiety concurrent with substance abuse disorder (marijuana and alcohol). He has been hospitalized and gone to rehabilitation multiple times. Now he’s adding RX abuse to his problems. I told him I was considering reporting him to his doctors, and he told me to keep my mouth shut and let him make his own decisions. I think his illness keeps him from making logical decisions about this, as he is constantly seeking ways to “not be bored”, “kill the anxiety”, or “not feel anything”.TLDR: SO says his decision to abuse prescription medications is not my problem, despite his diagnosed substance abuse disorders.[\colorbox{green}{\textbf{Interactive}}]&	\textit{Rant}	&	CR	&	\textit{Sadness}	&	\textit{Disappointment}	&	\textit{Average} \\
\cmidrule{1-6}
I live in Canada. I started the application process for medical assistance in dying. It was nice to finally receive an intelligible response to "I really want to die." as oppose to the usual parroted phrases and hollow cheerleading. Not sure that I will go through with it tbh, but I hope that I do. Getting two witness signatures seems like the only real obstacle here. [\colorbox{red!20}{\textbf{Isolated}}] & \textit{Chitchat}	&	CNR	&	\textit{Optimism}	&	\textit{Joy}	&	\textit{Excellent} \\
\cmidrule{1-6}
Who's an alcoholic stay at home wife/mom? Just super interesting to me. What are your days typically like? Do you hide your consumption or own it? [\colorbox{blue!20}{\textbf{Non-Interactive}}] & \textit{Survey}	&	NC	&	\textit{Excitement}	&	\textit{Neutral}	&	\textit{Excellent} \\
\cmidrule{1-6}
Every day I read to you from Daily Reflections. This book is published by Alcoholics Anonymous, and is an important resource for members of this LIFE-saving association.
Many recovering alcoholics use this literature to start their day. AA is a fellowship of men and women who share their strengths and hopes with each other for the common purpose of helping the alcoholic who still suffers.
It is the faith and the love of the LIFE which allows the cure or rather ashes the progression of the disease and helps us to leave the hell of alcoholism one day at the time. Addiction is very powerful and sneaky.
Although these writings are primarily intended for alcoholics, their families and friends, many people who feel they have no contact with alcoholism greatly appreciate the wisdom that emanates from them.
We hope you will gain the freedom we know!
 MERCI!
Thank you for your support by subscribing to our YouTube channel. <youtube channel>
\#recovery \#alcoholism \#alcoholics anonymous love \#LIFE \#January	[\colorbox{red!20}{\textbf{Isolated}}] &	\textit{Chitchat}	&	NC	&	\textit{Neutral}	&	\textit{Neutral}	&	\textit{Poor}\\

\bottomrule 
\multicolumn{6}{c}{} \\ 
\end{tabular}}
\caption{Example of posts and their corresponding labels in \peerData. \textbf{Intent:} \textit{Help-seeking}, \textit{Rant}, \textit{Chit-chat}, and \textit{Survey}; \textbf{Criticism:} \textit{Self-criticism} (SC), \textit{Criticism with reasoning} (CR), \textit{Criticism with no reasoning} (CNR), and \textit{No-criticism} (NC); \textbf{Readability:} \textit{Excellent} (5), \textit{Good} (4); \textit{Average} (3), \textit{Mediocre} (2), and \textit{Poor} (1); \textbf{Emotion:} \textit{Admiration, Amusement, Anger, Annoyance, Approval, Caring, Confusion, Curiosity, Desire, Disappointment, Disapproval, Disgust, Embarrassment, Excitement, Fear, Gratitude, Grief, Joy, Love, Nervousness, Optimism, Pride, Realization, Relief, Remorse, Sadness, Surprise}, and \textit{Neutral}; \textbf{Engagement:} \textit{Interactive, Non-interactive,} and \textit{Isolated}.
}
\label{tab:interaction_annot_example}
\end{table}

\section{Additional Auxiliary Analysis}
\label{sup:auxiliary}
This section analyses the auxiliary properties of the \peerData\ dataset. We run fundamental experiments to analyze \peerData's focus on metadata and textual properties.

\subsubsection*{Title Length in \peerData}
\label{title_length}
We run a short analysis on the length of the titles of posts from each category in \peerData. The aim is to comprehend if more descriptive titles are helpful while receiving help or make it cumbersome for helpers to skip over the post. We find that the mean length of the title for interactive posts is greater than that of non-interactive and isolated posts by a margin of $3.4$ and $5.5$ words, respectively. These statistics indicate that descriptive titles are more explanatory and likely to receive help. We further conclude that the original poster, who wrote descriptive titles, is likelier to engage with the peers trying to help. This explains the fact that peers primarily look for self-explanatory titles to decide to help.

\subsubsection*{Body Length in \peerData}
Once a descriptive title catches the helper's attention, we hypothesize that the body of the text should also be descriptive enough to impart proper information to receive help. To verify this, we run a simple analysis of the body lengths of the post. Furthermore, the statistics show that the mean length of the body for interactive posts ($189.95$) is greater than that for non-interactive and isolated posts by a margin of $25.64$ and $35.24$, respectively. This supports our hypothesis, indicating that help-seekers who write descriptive posts are more likely to receive help. We also analyze that the mean body length for non-interactive posts is greater than for isolated posts by a margin of $9.6$. Once again, this indicates that posts with longer body lengths are more likely to attract helpers by imparting information better.

\subsubsection*{The Number of Peer Comments}
Here, we analyze how many peers interact with the posts. On average, an interactive post receives $10.12$ comments, whereas a non-interactive post receives merely $3.41$ comments. Evidently, the number of peer comments on interactive posts is tripled compared to the peer comments on non-interactive posts by a margin of $6.71$. 
This analysis significantly differentiates the interactivity among the interactive and non-interactive posts on mental health subreddits. Moreover, the average number of unique peers interacting on the interactive posts is more than two times the number of non-interactive posts. The same trend holds for other statistical measures. For interactive and non-interactive posts, we observed a standard deviation of 12.47 and 5.94, median of 4 and 2, and 95th percentile of 22 and 10, respectively.

\subsubsection*{The Number of Unique Peers on a Thread}
\label{unique_peers}
The engagement category influences the number of peers participating in a conversational thread. 
Our findings state that an average interactive post attracts $6.07$ unique participants on the thread, and a non-interactive post receives merely $3.06$. This shows that the average number of unique peers on interactive posts is almost double that of peers interacting on non-interactive posts. The help-seeker interacts with the peers by writing replies to their initial replies. The result indicates that the peers are more interactive, and the number of new peers pitching in to help also increased. Consequently, the help-seeker's interactivity also increases the post's overall engagement, resulting in an increase in the chance of receiving assistance. 

\begin{figure}[!t]
\includegraphics[width=\textwidth]{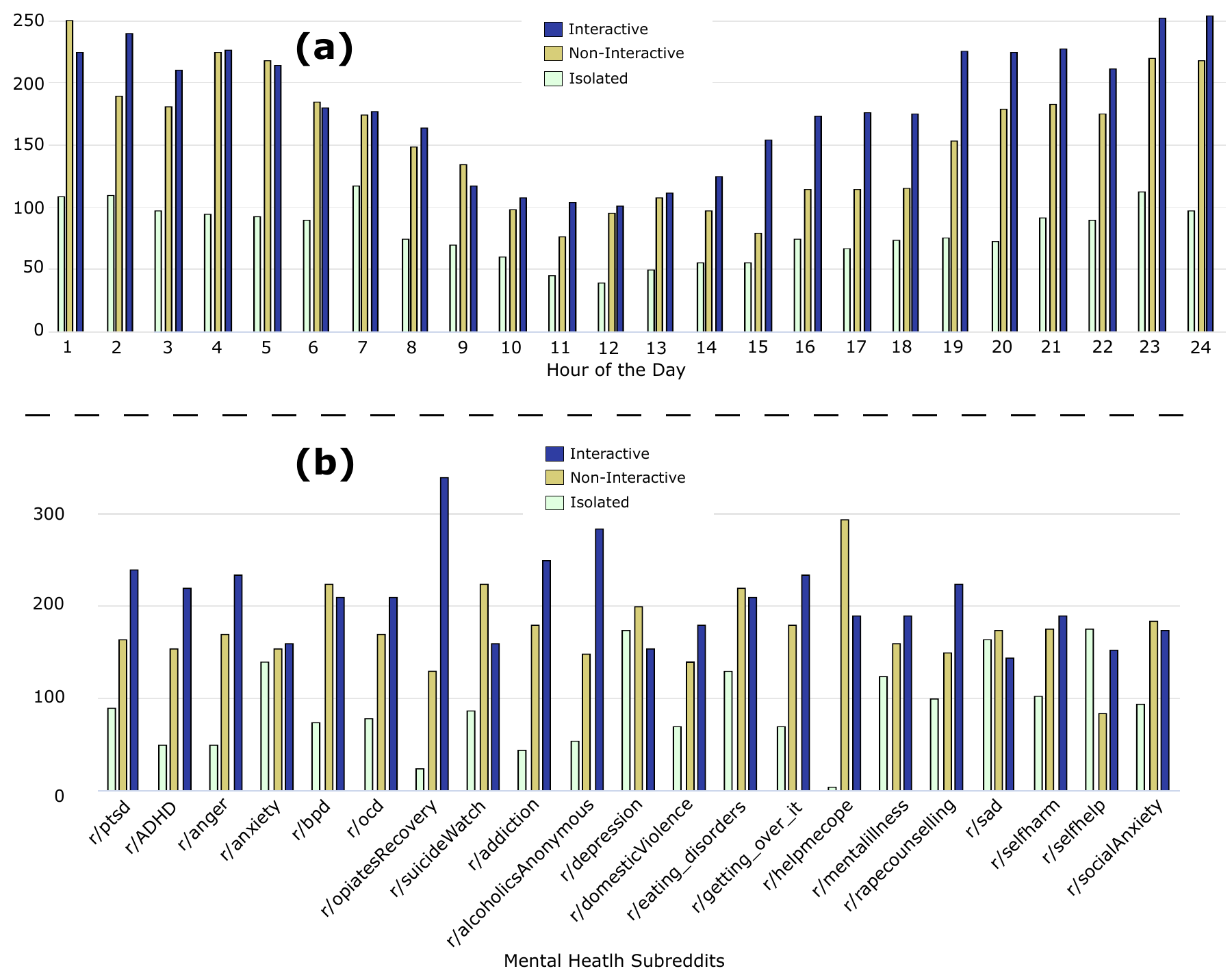}
\caption{{\bf (a)} The time distribution of posting mental health-related posts on OMHCs. The plotted timestamps record UTC timezone and demonstrate that day hours attract the least activity on the platform, whereas night hours attract the maximum activity. {\bf (b)} Distribution of mental health subreddits across all engagement categories.}
\label{engagement_time}
\end{figure}

\subsubsection*{Time of Posting (Seeking Help)}
Studies showed that users' social media presence is highly correlated with the time of the day\footnote{https://seopressor.com/blog/science-behind-best-worst-time-to-post-social-media/}. We hypothesize that both help-seekers and helpers find it hard to go online during working hours and seek or give help. We run a timestamp-based analysis to understand the frequency of posting during different times of the day. Figure \ref{engagement_time} shows that in all three categories, the lowest number of posts are made between 1000 and 1300 hours UTC. Also, 2400 hours UTC is when the maximum number of interactive posts arrived. Similarly, the maximum number of non-interactive posts arrived at 0100 to 0200 hours UTC, and the maximum number of isolated posts arrived from 0700 to 0800 hours UTC. Evidently, this supports our hypothesis that people suffering from mental health issues tend to seek help mostly at night. Subsequently, we could also recommend help-seekers to seek help during potential hours when help is most likely to be received.

\subsubsection*{Subreddit Division}
We assess which mental health issues are more prevalent in peer-to-peer counseling. Also, it is an interesting comparative study to analyze which mental health issues usually attract interaction. As shown in Figure 
\ref{engagement_time}, \textit{r/helpmecope} contributes to the minimum number of isolated posts out of all subreddits and the maximum number of non-interactive posts. The subreddit \textit{r/OpiatesRecovery} contributes to the maximum number of interactive posts to the dataset. The number of isolated posts is greater than that of non-interactive posts, which in turn was greater than the number of interactive posts for $13$ out of $21$ mental health subreddits in the \peerData\ dataset. 

\subsubsection*{Anonymity} 
Reddit allows users to create \textit{throwaway} accounts, which are temporary accounts for creating just an anonymous post. Thus, any content posted via an anonymous account does not reveal the user's personal account information. To explore the effect of anonymity on peer-to-peer counseling conversations, we extract all the posts with the throwaway user handles from \peerData. Next, we analyze these posts for differentiating characteristics. We find $245$ posts from throwaway accounts in \peerData. Furthermore, $28$ out of $245$ posts are isolated, $101$ are non-interactive, and $116$ are interactive. Throwaway accounts contribute to a meagre $2.4\%$ out of total posts. This indicates that help-seekers barely use the added anonymity offered by the OMHC platforms for posting mental health issues. 
Since it is clear that the posts are less likely to receive a reply if posted from throwaway accounts, we explore the alternate possibility of checking if the help varies with anonymity (Case 1). We also verify if peers' behavior while replying changes when they move from a regular to an anonymous account (Case 2). 
To this, we test the following two cases:

\noindent\textbf{Case 1:} Posts made from throwaway accounts are less likely to receive help as compared to the posts made from any regular account.

\noindent\textbf{Case 2:} Help-seekers posting from throwaway accounts are more likely to reply on receiving replies than those posting from regular accounts.

We observe that $88.5\%$ posts by throwaway accounts received a reply compared to $81.3\%$ posts that were not from throwaway accounts. Thus, the finding does not support our hypothesis. Moreover, the original poster replied in $53.5\%$ of the posts from throwaway accounts, which received at least a reply. This is similar to the $53.8\%$ posts not from throwaway accounts. Consequently, we do not observe any correlation between the behavior of the help seeker and anonymity; thus, the second case also contradicts the findings.

\subsubsection*{Analysis Based on Similarity}
Earlier work showed that better responses from counselors uphold higher linguistic similarity with the clients’ responses. Thus, we test the hypothesis to observe if the interactive engagements carry a higher similarity between the post and the comments than other engagements. 
Hence, to shed light on this and further clarify, we exploit sentence-bert \cite{reimers-2019-sentence-bert} to compute the cosine-similarity between the context-rich BERT embeddings of help-seeker's post and peer comments. As a result, we observe that the earlier work's hypothesis contradicts the findings of our analysis. The average text-similarity between posts and comments for non-interactive posts is very similar to that of interactive posts with a slight difference of $0.0202$ only. Thus unlike earlier work, the high textual similarity between the post and the comments is not a standard case.

\section{Discussion}

The process of data preparation for behavioral analysis in mental health is a critical step in itself. This is due to the multifaceted nature of how individuals express themselves and interact within these contexts. In the BeCOPE dataset, Selecting suitable labels was a careful process that involved the assistance of mental health experts. A team of experts meticulously analyzed a sample of posts, aiming to grasp the overarching posting behaviors demonstrated by peers within mental health-specific subreddits. This analysis was complemented by a thorough review of prior studies in a similar space. Despite the existence of works in understanding OMHC platforms, our work underscores the utmost relevance toward the need for such behavioral analysis while simultaneously presenting a novel perspective compared to the existing works.

We discussed a sequence of relevant analyses, supplementary to the primary analysis detailed in the the main text. Notably, intrinsic characteristics exhibited by help-seekers significantly influence their engagement dynamics and distinctly bifurcate those receiving interaction from those not. For instance, a recurrent behavioral pattern observed throughout the dataset is the tendency of individuals to post during late hours, presumably when they find spare time from their busy daily schedules. Furthermore, Reddit offers an exclusive feature enabling users to interact incognito (going anonymous while posting). Given the prevailing stigma associated with open discussions in this domain, a prevailing assumption was that engagement in anonymous interactions would surpass non-anonymous posts. Contrarily, our analysis disproved this expectation, revealing that anonymous posts actually attract lower levels of interaction. The crux of our research underscores the need for widespread education regarding the optimal "what, when, and how" of engaging on OMHCs. Such knowledge spread holds the potential to enhance the productive utilization of OMHC platforms and directly impact the efficacy of peer-to-peer support provision among individuals seeking help.


\end{document}